\documentclass[sigconf]{acmart}
\usepackage{multirow}
\AtBeginDocument{%
  }

\setcopyright{acmlicensed}
\copyrightyear{2026}
\acmYear{2026}
\acmDOI{XXXXXXX.XXXXXXX}
\acmConference[Conference acronym 'XX]{Make sure to enter the correct
  conference title from your rights confirmation email}{June 03--05,
  2018}{Woodstock, NY}
\acmISBN{978-1-4503-XXXX-X/2018/06}

\begin{document}

\title{KMLP: A Scalable Hybrid Architecture for Web-Scale Tabular Data Modeling}

\author{Mingming Zhang}
\authornotemark[1]
\orcid{0009-0008-0434-5053}
\affiliation{%
  \institution{Zhejiang University}
  \city{Hangzhou}
  \country{China}
}
\affiliation{%
  \institution{Ant Group}
  \city{Hangzhou}
  \country{China}
}
\email{mmz@zju.edu.cn}

\author{Pengfei Shi}
\authornote{Both authors contributed equally to this research.}
\orcid{0009-0006-0156-0079}
\affiliation{%
  \institution{Ant Group}
 \city{Hangzhou}
  \country{China}
}
\email{lasou.spf@antgroup.com}

\author{Junbo Zhao}
\orcid{0000-0002-3637-2936}
\affiliation{%
  \institution{Zhejiang University}
  \city{Hangzhou}
  \country{China}}
\email{j.zhao@zju.edu.cn}

\author{Ningtao Wang}
\orcid{0009-0005-6577-5047}
\authornote{Ningtao Wang is the corresponding author.}
\affiliation{%
  \institution{Ant Group}
 \city{Hangzhou}
  \country{China}
}
\email{ningtao.nt@antgroup.com}

\author{Feng Zhao}
\orcid{0009-0001-2239-2071}
\affiliation{%
  \institution{Ant Group}
 \city{Hangzhou}
  \country{China}
}
\email{zhaofeng.zhf@antgroup.com}

\author{Guandong Sun}
\orcid{0009-0007-4882-9183}
\affiliation{%
  \institution{Ant Group}
 \city{Hangzhou}
  \country{China}
}
\email{guandong.sgd@antgroup.com}

\author{Yulin Kang}
\orcid{0000-0002-8150-4854}
\affiliation{%
  \institution{Ant Group}
 \city{Hangzhou}
  \country{China}
}
\email{yulin.kyl@antgroup.com}

\author{Xing Fu}
\orcid{0000-0002-3536-2779}
\affiliation{%
  \institution{Ant Group}
 \city{Hangzhou}
  \country{China}
}
\email{zicai.fx@antgroup.com}

\author{Zhiqing Xiao}
\orcid{0009-0007-4889-644X}
\affiliation{%
  \institution{Zhejiang University}
 \city{Hangzhou}
  \country{China}
}
\email{zhiqing.xiao@zju.edu.cn}

\author{Weiqiang Wang}
\orcid{0000-0002-6159-619X}
\affiliation{%
  \institution{Ant Group}
 \city{Hangzhou}
  \country{China}
}
\email{weiqiang.wwq@antgroup.com}

\author{Ruizhe Gao}
\orcid{0009-0007-3690-3147}
\affiliation{%
  \institution{Ant Group}
 \city{Hangzhou}
  \country{China}
}
\email{ruizhe.grz@antgroup.com}

\renewcommand{\shortauthors}{Mingming Zhang et al.}

\begin{abstract}
Predictive modeling on web-scale tabular data presents significant scalability challenges for industrial applications, often involving billions of instances and hundreds of heterogeneous numerical features. The inherent complexities of these features—characterized by anisotropy, heavy-tailed distributions, and non-stationarity—not only impose bottlenecks on the training efficiency and scalability of mainstream models like Gradient Boosting Decision Trees (GBDTs), but also compel practitioners into laborious, inefficient, and expert-dependent manual feature engineering.
To systematically address this challenge, we introduce KMLP, a novel hybrid deep architecture. KMLP synergistically integrates a shallow Kolmogorov-Arnold Network (KAN) as a front-end with a Gated Multilayer Perceptron (gMLP) as the backbone. The KAN front-end leverages its learnable activation functions to automatically model complex non-linear transformations for each input feature in an end-to-end manner, thereby automating feature representation learning. Subsequently, the gMLP backbone efficiently captures high-order interactions among these refined representations.
Extensive experiments on multiple public benchmarks and an ultra-large-scale industrial web dataset with billions of samples demonstrate that KMLP achieves state-of-the-art (SOTA) performance. Crucially, our findings reveal that KMLP's performance advantage over strong baselines like GBDTs becomes more pronounced as the data scale increases. This validates KMLP as a scalable and adaptive deep learning paradigm, offering a promising path forward for modeling large-scale, dynamic web tabular data.
\end{abstract}


% View CCS Display


% <concept_id>10003033.10003034</concept_id>
% <concept_desc>Networks~Network architectures</concept_desc>
% <concept_id>10010405.10010406.10010424</concept_id>
% <concept_desc>Applied computing~Enterprise modeling</concept_desc>
% <concept_id>10010147.10010257</concept_id>
% <concept_desc>Computing methodologies~Machine learning</concept_desc>

% \ccsdesc[500]{Networks~Network architectures}
% \ccsdesc[500]{Applied computing~Enterprise modeling}
% \ccsdesc[500]{Computing methodologies~Machine learning}



\maketitle

\section{Introduction}
Predictive modeling of user behavior is a cornerstone of modern web platforms, from e-commerce and social networks to online financial services. These platforms generate billions of user interaction events daily, forming web-scale tabular data with hundreds of heterogeneous features. Unlike the static datasets found in traditional benchmarks, this real-world web data exhibits significant dynamism: its feature distributions (e.g., user spending habits, click-through patterns) are complex and evolving, often accompanied by challenges such as anisotropy and heavy-tailed distributions. This dynamic nature renders traditional modeling paradigms that rely on manual feature engineering unsustainable, as expert-crafted rules struggle to maintain long-term effectiveness in a rapidly changing web environment.

Traditional ensemble methods based on decision trees, such as Gradient Boosting Decision Trees (GBDT) \cite{chen2016xgboost, ke2017lightgbm, dorogush2018catboost}, have long dominated the field of tabular data processing. These models are widely adopted due to their nonparametric nature, which makes no assumptions about training data or prediction residuals. Moreover, tree-based models effectively capture complex nonlinear relationships among heterogeneous variables in tabular data without requiring extensive preprocessing of feature columns. However, when confronted with web-scale data, their advantages begin to be eroded by inherent limitations. First, on the front of scalability and efficiency, the sequential training mechanism of GBDTs hinders their ability to fully leverage modern distributed computing resources. For web datasets comprising billions of samples, this leads to prohibitive training times and high computational costs, severely impeding the rapid model iteration and A/B testing essential for web applications. Second, a more fundamental limitation lies in their static learning paradigm. GBDTs construct decision trees using fixed split points in a "one-shot" learning process. When handling numerical features, this mechanism essentially discretizes the continuous feature space into a fixed set of bins. As the feature distributions in web data evolve (e.g., a general upward shift in the range of users' average spending), the original split points may no longer be optimal, thereby limiting the model's expressive power. The only recourse to capture these new patterns is to rely on a new round of time-consuming manual feature engineering and initiate costly full model retraining.

Although various deep learning methods, such as TabNet \cite{arik2019TabNet}, SAINT \cite{somepalli2021saint} and TabTransformer \cite{huang2020tabtransformer}, have emerged in recent years and demonstrated the potential to outperform traditional tree-based models on smaller publicly available datasets, their scalability and robustness in Web-scale industrial settings remain underexplored.

However, our findings suggest that the performance of neural network models can be substantially improved, potentially surpassing that of tree-based models, by incorporating expert-designed features and business-specific feature engineering. For instance, in the context of loan businesses, features derived from users' spending amounts on specific platforms, aggregated using time-window operations (e.g., calculating the maximum and mean), can improve predictive accuracy. Nevertheless, the development of such sophisticated features typically relies on the expertise of domain specialists and iterative experimental design.
This frames a critical \textit{web mining research question}: \textbf{Can we design a scalable and adaptive deep learning architecture that can automatically learn effective feature representations from dynamic, complex web data, thereby obviating the heavy reliance on manual feature engineering and frequent retraining?}

To address this challenge, we propose KMLP, a novel hybrid deep architecture tailored for web-scale tabular data. The core idea of KMLP is to decouple automated feature transformation from high-order interaction modeling. It employs a shallow Kolmogorov-Arnold Network \cite{liu2024kan} as a front-end, which leverages its learnable activation functions to learn a continuous and adaptive non-linear transformation for each raw web feature (e.g., time since last login, total historical transaction value) in an end-to-end fashion. This process not only automates traditional manual feature engineering but, more importantly, can flexibly respond to shifts in feature distributions without being constrained by the fixed split points inherent to GBDTs. Subsequently, the refined feature representations from the KAN are fed into a Gated Multilayer Perceptron (gMLP) backbone, which efficiently captures high-order interactions among these powerful representations.
We conducted extensive experiments on multiple public benchmarks and a real-world lending dataset from a global financial web platform, comprising billions of samples. The results demonstrate that KMLP achieves state-of-the-art (SOTA) performance across all tests. Crucially, our findings reveal that KMLP's performance advantage over strong baselines like GBDTs becomes more pronounced as the data scale increases. This validates KMLP as a scalable and adaptive deep learning paradigm, offering a promising path forward for modeling dynamic web tabular data.

The main contributions of our paper are as follows:
\begin{itemize}
\item \textbf{Hybrid Architecture and Methodology}:
We propose KMLP, the first work to leverage a shallow KAN as an adaptive feature engineering constructor, combined with gMLP for efficient interaction modeling, enabling high-quality tabular feature representations. Systematic evaluation shows that applying a Quantile Linear Transformation (QTL) mitigates feature heterogeneity and stabilizes training, further enhancing KMLP’s robustness and performance on large-scale web tabular data.

\item \textbf{Scalability and Data-Scale Advantage}:
The scalability of KMLP is evaluated on large-scale web tabular data, including a real-world financial and payment platform dataset with billions of records.
As data volume grows, KMLP outperforms traditional tree-based models, showing that deep learning’s advantage on tabular data arises from both model design and data scale. This is the first systematic study highlighting the critical role of data scale in tabular neural networks.
 \item \textbf{SOTA Performance on Web-Scale Financial Data}: 
 KMLP achieves SOTA performance on public tabular benchmarks and a 1.76-point KS improvement on industrial-scale web data, demonstrating robustness and adaptability to massive, heterogeneous environments. In real-world financial web deployments, it cut overdue loans by USD 46 million and increased credit scale by USD 670 million within six months. 
\end{itemize}
\section{Related Works}

\subsection{Table Prediction}

In the important area of tabular prediction, deep learning algorithms have been increasingly integrated with traditional methods, significantly advancing the progress of this field.

\noindent
\textbf{Traditional Methods.} play a crucial role in supervised and semi-supervised learning, particularly with tabular datasets. Among these methods, tree-based models have become the leading choice. Widely used tools like XGBoost \cite{chen2016xgboost}, CatBoost \cite{dorogush2018catboost}, and LightGBM \cite{ke2017lightgbm} have been successfully applied in various practical scenarios due to several key advantages. These advantages include strong interpretability, the ability to easily handle different types of features (including missing values), and exceptional performance across various data scales. However, when the scale of tabular data is extremely large, tree models often encounter problems such as excessive consumption of computing resources and overly long training times.
\noindent
\textbf{Neural Networks.} In recent years, with the rapid advancement of deep learning, numerous deep learning models have been progressively applied to the domain of tabular data processing. For instance, TabNet \cite{arik2019TabNet} innovatively leverages neural networks to emulate decision trees, focusing selectively on a small number of key features at each layer to emphasize their importance. NODE  \cite{Popov2020Node}  combines neural oblivious decision trees  with dense connections, achieving performance comparable to gradient-boosted decision trees.
With the increasing popularity of attention mechanisms, several Transformer-based models for tabular data have also been proposed. For example, TabTransformer \cite{huang2020tabtransformer} primarily utilizes a Transformer encoder to generate contextual embeddings for categorical features. In contrast, FT-Transformer \cite{gorishniy2021fttransfomers}  maps both continuous and categorical features into a higher-dimensional embedding space before processing them through Transformer blocks, utilizing attention mechanisms to generate more effective feature representations. The SAINT \cite{somepalli2021saint} model focuses on the row and column attention of the table. It captures the association information between rows and columns through a unique attention mechanism, bringing a new perspective to the understanding and analysis of tabular data.

On Web-scale datasets, these methods often fail to deliver the expected performance and may even underperform compared to tree-based models. However, while tree models excel on small- to medium-sized tabular data, they encounter significant challenges in large-scale datasets, particularly in terms of computational resource requirements and time efficiency. As a result, we are motivated to further explore neural network-based solutions to develop more efficient and scalable approaches to address these challenges.

\subsection{Kolmogorov-Arnold Networks}
% kan的介绍
% TODO: Kolmogorov - Arnold  -> Kolmogorov-Arnold ?
The Kolmogorov-Arnold Networks \cite{liu2024kan} has garnered significant interest in the field of neural network architectures. Its foundation is based on the Kolmogorov-Arnold approximation theorem \cite{HechtNielsen1987KolmogorovsMN}, which serves as a theoretical cornerstone for the emergence and advancement of KAN. According to this theorem, multivariate continuous functions can be expressed as a finite sum of univariate continuous functions. In formula, for a $d$-dimensional continuous function $f:[0,1]^d\rightarrow \mathbb{R}$, it can be represented as,
\begin{equation}
    f(x_1,x_2,...,x_d) = \sum_{q=1}^{2d+1}\Phi_q\left(\sum_{p=1}^{d}\phi_{q,p}(x_p)\right),
\end{equation}
where each function $\phi_{q,p}:[0,1] \rightarrow \mathbb{R}$ and $\Phi_q: \mathbb{R} \rightarrow \mathbb{R}$ are continuous.
% Inspired by representation theorem,  a Kolmogorov-Arnold layer learn univariate functions.  A general KAN network is a stacking of $L$ Kolmogorov-Arnold layers. The characteristics of learnable activation functions in KAN allow us to capture the complex data patterns by only using shallow layers.
% call back motivation
Inspired by the representation theorem, Kolmogorov-Arnold layers are capable of learning univariate functions. A general Kolmogorov-Arnold Network (KAN) is constructed by stacking \( L \) Kolmogorov-Arnold layers. The key advantage of KAN lies in its learnable activation functions, enabling the network to effectively capture complex data patterns even with a shallow architecture. This characteristic provides significant motivation for replacing traditional handcrafted feature engineering.

Traditional feature engineering heavily relies on domain expertise and manual efforts, which not only make the process time-consuming but also limit its ability to fully uncover complex and hidden patterns within the data. In contrast, shallow KANs leverage their powerful representation learning capabilities to automatically extract relevant features directly from raw data, eliminating the need for labor-intensive manual design. Moreover, the efficient feature extraction capability of shallow KANs can identify hidden data structures that might be overlooked by conventional methods, thereby improving task performance. This not only simplifies the data processing workflow but also unlocks new potential for discovering intricate and sophisticated data patterns.

\subsection{gMLP}
gMLP \cite{gmlp} is an innovative enhancement of the traditional MLP. By cleverly introducing a gated mechanism, it significantly improves the model's data processing capability and adaptability.
% 加公式描述
 
SwiGLU \cite{Shazeer2020GLUVI} is a commonly used activation function in gMLP. It functions within each neuron or layer, helping gMLP to perform nonlinear transformations on the input data, thereby enhancing the expressive power of the model. 
The gMLP model has demonstrated competitive performance in both the language and vision domains. For instance,
Google's PaLM \cite{Chowdhery2022PaLMSL} and Meta's LLaMA \cite{Touvron2023LLaMAOA} utilize SwiGLU to enhance the performance of the FFN layers in the Transformer architecture.

% 说明一下为什么我们的模型要用gmlp？
Compared with the traditional MLP, gMLP's unique gated mechanism enabled it to better manage the sequential nature of language data, resulting in higher accuracy and faster convergence rates when compared to traditional MLPs.

\section{METHODOLOGY}

% 插入模型
\begin{figure*}[h]
  \centering
  \includegraphics[width=\textwidth]{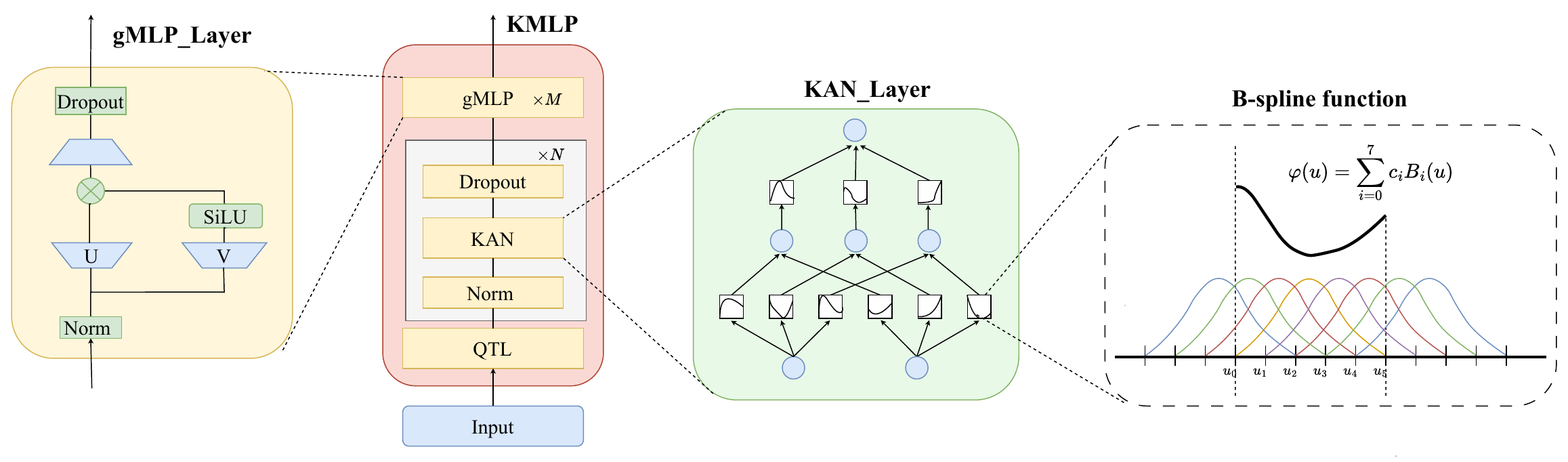}
  \caption{
  \small
  Overview of the KMLP-QTL structure. Tabular data features are first processed by QTL for fine-grained numerical representation. The data then flows through the KAN layer to manage feature heterogeneity and complex interactions, followed by stacked gMLP modules to capture deep non-linear interactions. Batch Normalization and Dropout are included in intermediate layers to enhance performance and stability. }
  \label{fig:model}
\end{figure*}

%1. model 
%2. preprocessing
%   
Given that real-world industrial datasets are predominantly composed of continuous features, we first propose a nonparametric preprocessing method specifically designed for numerical features. Subsequently, we will introduce the hybrid architecture combining KAN and gMLP in detail. The overall framework structure we propose is illustrated in Figure \ref{fig:model}. 

\subsection{Preprocessing Method for Numerical Features}
The performance of neural network algorithms can be significantly influenced by the scale of the data. As a result, it is essential to normalize \cite{ali2014data, singh2020investigating, AKSOY2001563} both the input data and the intermediate layers of the neural network to enhance the model's learning ability. 
For a supervised learning task on tabular data, we can represent the dataset as $\{(x_j,y_j)\}_{j=1}^{m}$. Here, $y_j$ denotes the label, and $x_j=\left(x_j^{(\text{num})},x_j^{(\text{cat})}\right)$ represents the features of the object, which include both numerical and categorical attributes.
For categorical features, one hot encoding can be adopted.
If the tabular data contains categorical features, we can use one-hot encoding or apply the lookup operation to map them into a learnable embedding.
Since the industrial tabular datasets are mainly composed of numerical features, we mainly discuss the processing methods for numerical features. Previous related work \cite{guo21ctr,gorishniy2022on} has pointed out that changing the representation form of the original scalar values of numerical features can improve the learning ability of tabular deep learning models.
In a formula, we need to find numerical transform as,
\[
z_j = f_j\left(x_j^{(\text{num})}\right).
\]
Here $f_j$ is the embedding function for the $j$-th numerical
feature, $z_j$ is the embedding of the $j$-th numerical feature.
\begin{figure}[h]
  \centering
  \includegraphics[width=\linewidth]{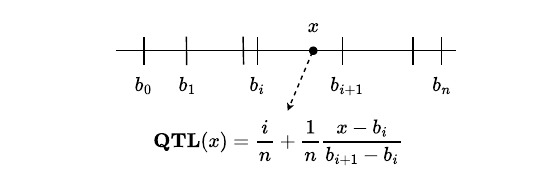}
  \caption{
  \small
  Quantile Transformation with Linear interpolation }
  \label{fig:QTL}
\end{figure}

In practical industrial applications, discretizing continuous numerical features into categorical features through binning operations is a widely adopted approach. However, this binning method has inherent limitations: it fails to capture the magnitude of values within individual bins as well as the relative relationships between values across bins. To address these shortcomings, we propose an improved discretization strategy. Specifically, we map the distribution of the original data to an approximately uniform distribution. For the $j-$th numerical feature, we divide its value into $n$ disjoint intervals $(b_{i},b_{i+1}]$, where $b_{i}$ is the $i$th-quantile. We utilize \textbf{Q}uantile \textbf{T}ransformation with \textbf{L}inear interpolation as :
% Defined the scheme of the quantile linear encoding as:
\begin{equation}
\label{eq:QTL}
   \textbf{QTL}(x) = \frac{i}{n}+\frac{1}{n}\frac{x-b_{i}}{b_{i+1}-b_i}.
\end{equation}

QTL achieves global distribution uniformity while preserving the original intra-bin order of the data, thereby effectively avoiding extreme information distortion caused by discretization. This approach is particularly suited for scenarios where the relative magnitude of numerical features plays a critical role in prediction outcomes (e.g., user behavioral features in credit default prediction). 

There is a newly developed method, the piecewise linear encoding (PLE)~\cite{gorishniy2022on}, for numerical embedding that is similar to our nonparametric ideas, 

\begin{equation}\label{eq:ple}
\begin{split}
\text{PLE}(x)& = [e_1,...,e_n]\in \mathbb{R}^{n}\\
\quad\text{where~} e_i &=
\begin{cases}
    0, & x<b_i,\\
    1, & x\geq b_{i+1},\\
    \frac{x-b_{i}}{b_{i+1}-b_{i}}, & \text{otherwise.} 
\end{cases}
\end{split}
\end{equation}
We have noticed that it transforms each one-dimensional feature into an n-dimensional embedding, which consequently increases the input dimension of the model significantly and may lead to the problem of both dimensionality curse and interpretability. In the experimental section, we will also compare the performance of this operator and our proposed operator on tabular tasks.

\subsection{KMLP}
As illustrated in Figure \ref{fig:model}, KMLP (Kolmogorov-Arnold Network with gated MLP) presents a novel hybrid architecture specifically designed to address the unique challenges of tabular data representation and prediction. Tabular data is characterized by heterogeneous features, including numerical variables with vastly different statistical distributions and categorical variables with varying cardinalities. These properties make tabular data notoriously difficult for neural network-based models to process effectively, often allowing traditional tree-based models to outperform state-of-the-art architectures like Transformers.

KMLP aims to address this gap by leveraging the complementary strengths of KAN (Kolmogorov-Arnold Network) and gMLP (gated Multi-Layer Perceptron). The design philosophy underpinning KMLP is twofold: first, model feature heterogeneity through adaptive learnable transformations (KAN); second, capture complex homogeneous feature interactions and patterns through lightweight and computationally efficient mechanisms (gMLP). This hybrid structure ensures that KMLP can overcome both the heterogeneity and interaction modeling challenges in tabular data to deliver superior predictive performance without sacrificing computational scalability. KMLP resolves these challenges by combining:

(1) \textbf{KAN}:
Designed for diverse raw tabular features, KAN introduces adaptive activation functions capable of learning the inherent mappings between heterogeneous features. These transformations reduce reliance on manual feature engineering by domain experts, enabling the model to automatically generate optimized feature embeddings.
For the KAN layer,
\begin{equation}
    \textbf{KAN}(\textbf{x}) = \sum_{q=1}^{2d+1}\Phi_q\left(\sum_{p=1}^{d}\phi_{q,p}(x_p)\right),
\end{equation}
here, $\phi_{q,p}$ is referred to as the inner function, and $\Phi_q$ is referred to as the outer  function. They can be expressed in the form of a linear combination and B-spline functions as follows:
\begin{equation}
\label{bx}
    \varphi(x)=w_b\frac{x}{1+e^{-x}}+w_s\sum c_iB_i(x),
\end{equation}
where, $B_i(x)$ is a B-spline function, $w_b$ and $w_s$ are weight parameters, and $c_i$ is a control coefficient for shaping the B-spline.
% 增加一段B-spline的描述：
B-splines are widely used in computer-aided design, computer graphics, and numerical analysis to represent curves and surfaces. A B-spline curve of degree is defined as a linear combination of control points and basis functions.
The basis functions of  B - splines are defined over a knot vector. Let $U=\{u_0,u_1,...,u_m\}$ be a non-decreasing sequence of real numbers called the knot vector,  the $i$-th B-spline basis function of degree $p$ , denoted as $N_{i,p}(u)$ , can be recursively defined as follows:
\[N_{i,p}(u)=
\begin{cases}
    I\{u_i\leq u \leq u_{i+1}\}, &p=0,\\
    \frac{u-u_i}{u_{i+p}-u_i}N_{i,p-1}(u)+\frac{u_{i+p+1}-u}{u_{i+p+1}-u_{i+1}}N_{i+1,p-1}(u), &p>0,
\end{cases}
\]
where $I$ is an indicator function. The B-spline of 3 degrees (p=3) is employed, represented as \( B_i = N_{i,3} \) in Equation  (\ref{bx}).

In our experiments, we employed the cubic B-spline. We utilized Efficient-KAN (for code implementation, refer to \cite{Blealtan2024EfficientKan}), a reformulation of the originally proposed KAN method \cite{liu2024kan}, which significantly reduces memory consumption and improves computational efficiency.

% The B-spline of 3 degrees (p=3) is employed,  We use Efficient-KAN (for code implementation
% refer to \cite{Blealtan2024EfficientKan}), a reformulation of originally
% proposed KAN \cite{liu2024kan} which significantly reduces the memory
% cost and make the computation faster.

(2)  \textbf{gMLP}:
Once feature heterogeneity is addressed in the shallow KAN layers, gMLP operates on the harmonized embeddings to capture non-linear feature interactions and relationships. These lightweight gated modules ensure computational efficiency while providing the expressive capability required for high-quality tabular data predictions.
For the gMLP block,
\begin{align}
% \label{eq:gmlp}
    \textbf{x} &= \textbf{Batch Norm}(\textbf{x}) \label{eq:bn}\\
    \textbf{x} &= \textbf{SwiGLU}(\textbf{x})= \textbf{SiLU}(\textbf{x}V+b_1)\otimes (\textbf{x}U+b_2) \label{eq:glu}\\
    \textbf{x} &= \textbf{Dropout}(\textbf{x}) \label{eq:dp}
\end{align}
Here, $\textbf{SiLU}$ is sigmoid linear unit, defined as 
\(\textbf{SiLU}(x)=x\cdot\textbf{Sigmoid}(x)\).
% \begin{equation}
% \textbf{SiLU}(x)=x\cdot\textbf{Sigmoid}(x),
% \end{equation}
$U$ and $V$  represent linear mappings in different channels, respectively, and $b_i$ represents the corresponding bias. $\otimes$ represents multiplying element by element correspondingly.

\section{Experimental Setup}
\subsection{Dataset}
The experimental dataset consists of two parts as follows:

(1) \textbf{Open Public Datasets}: We selected six public tabular classification datasets from the OpenML platform. For
each dataset, we form 70\%/10\%/20\% train/validation/test
splits, where a different split is generated for every trial and all methods use the same splits.  The
datasets include: 
% CP
Click\_prediction\_small (CP, Advertisement Click Prediction \cite{kddcup2012-track2});
% MT
MagicTelescope (MT, Detection of high-energy gamma particles \cite{MT};
% CD
Credit (CD, Bank Credit Decision \cite{CD});
% EG
Eeg-eye-state (EG, Health and Medicine \cite{eeg_eye_state_264});
% HI
Higgs (HI, simulated physical particles \cite{HI});
% JA
Jannis (JA, anonymized dataset \cite{JA}). The detailed information about the tables is provided in the appendix \ref{appd:dataset}.

(2) \textbf{Ultra Large-Scale Web Dataset}: 
% We collect a credit scoring dataset provided by a world-leading online payment platform.
We evaluate our model on a real-world, billion-scale credit scoring dataset from a world-leading online financial platform. This dataset captures the critical task of assessing credit risk for online consumer loans.
%Our research uses a dataset based on actual credit scoring situations, specifically related to loan delinquency. 
In this dataset, each record is labeled as "1" when the overdue period exceeds four months, thereby forming a binary classification dataset. The dataset consists of 449 numerical features, providing a comprehensive basis for analysis.
We divide our adopted dataset into the training set, the validation set and the test set according to the chronological order. To clearly illustrate the patterns of our model with respect to the scale of the training data, we present training datasets of different scales. The detailed information regarding the credit scoring dataset is shown in Table \ref{tab:ind_data}.

\begin{table}[t]
    \centering
    \caption{
    \small
    Statistical Information of the credit scoring dataset. It is a binary classification dataset. We prepared five training and validation sets with different data scales, while the test set is the same, with a proportion of 0.47\% for overdues (label=1) in the test set.}
    \label{tab:ind_data}
    \begin{tabular}{l c c c}
    \toprule
    Dataset  &\#Train &\#Valid &\#Test   \\ \midrule
    Tiny     &  200K & 50K  & 1B                  \\ \hline
    Small    &  2M   & 500K & 1B               \\ \hline
    Medium   &  20M  & 5M   & 1B              \\ \hline
    Large    &  200M & 50M  & 1B                \\ \hline
    Ultra Large     &  2B   & 50M  & 1B         \\
    \bottomrule
    \end{tabular}   
\end{table}

\begin{table*}[t]
    \centering
    \caption{
    \small
    Comparsion of different numerical preprocessing operators in open public datasets. We highlight the best results in bold. QTL achieved the best performance on 5
    out of 6 datasets.}
    \label{tab:num_encoding_openml}   
    \begin{tabular}{c c|c c|c c|c c|c c|c c|c c}
    \hline 
         \multicolumn{2}{c|}{\multirow{2}{*}{Operator}} & \multicolumn{2}{c|}{CP} &
         \multicolumn{2}{c|}{MT} &
         \multicolumn{2}{c|}{CD} &
         \multicolumn{2}{c|}{EG} &
         \multicolumn{2}{c|}{HI} &
         \multicolumn{2}{c}{JA} 
         \\
         \cline{3-14}
         \multicolumn{2}{c|}{} & AUC & KS &  AUC & KS &  AUC & KS  &  AUC & KS &  AUC & KS  &  AUC & KS \\ \hline
         \multicolumn{2}{c|}{CLR}    &69.18	&28.68	&93.35	&72.15	&85.16	&55.18	&81.84	&48.74	&80.03	&44.49	&87.03 &58.66\\ 
         \multicolumn{2}{c|}{Quantile} &64.22	&21.58	&92.53	&69.99	&82.37	&51.26	&64.65	&19.56	&79.38	&44.17	&86.67	&57.92 \\ 
         \multicolumn{2}{c|}{PLE}      &\textbf{69.20}	&\textbf{28.92}	&90.98	&67.43	&84.65	&54.15	&93.73	&72.46	&79.39	&43.47	&85.23	&55.00 \\ 
         \multicolumn{2}{c|}{\textbf{QTL}} &69.10	&27.76	&\textbf{93.72}	&\textbf{73.68}	&\textbf{85.20}	&\textbf{55.93}	&\textbf{99.16}	&\textbf{91.87}	&\textbf{80.88}	&\textbf{46.23}	&\textbf{87.30}	&\textbf{59.51}      \\ \hline
    \end{tabular}
\end{table*}

\subsection{Baselines}
In this subsection, we present various methods for preprocessing numerical features and comparison methods for models used in tabular prediction.

\noindent
\textbf{Preprocessing Baselines.}
We conduct experiments on numerical preprocessing methods to show the efficacy of QTL on tabular learning. 
\begin{itemize}
\item \text{Centered Log Ratio (CLR)}~\cite{FILZMOSER20104230} is a widely used preprocessing method that normalized the data under log-scale. Let $\textbf{x} = (x_1,x_2,...,x_d)$, 
\begin{equation}
    \text{CLR}(x_j) = \ln\left(\frac{x_j}{g(\textbf{x})}\right),
\end{equation}
where $g(\textbf{x}) = \left(\prod_{k=1}^{d}x_k\right)^{1/d}$.
\item \text{Quantile transformation} is an advanced Binning strategy that maintains the numerical relationships among bins. For each numerical feature $x$, we split its value into $n$ bins $(b_{i},b_{i+1}]$ for equal-frequency, the preprocessing operator is:
\begin{equation}
\text{Quantile}(x) = \frac{i}{n}. 
\end{equation}
\item \text{PLE}~\cite{gorishniy2022on} maps each individual feature into a $n$-dimensional embedding (see Equation (\ref{eq:ple})).
\end{itemize}

\noindent
\textbf{Architecture Baselines.}
We conduct experiments on the LightGBM and neural network-based methods to show the efficacy of KMLP on tabular learning. 

\begin{itemize}
\item \text{LightGBM}~\cite{ke2017lightgbm} is an efficient gradient-boosting framework-based machine learning algorithm that uses tree learning for rapid training and supports high-dimensional sparse data.
\item \text{MLP}~\cite{mlp} is a feedforward artificial neural network composed of multiple fully connected layers, with each layer's output serving as the next layer's input.
\item \text{gMLP}~\cite{gmlp} is a neural network architecture that incorporates gating mechanisms to enhance the model's expressive power by controlling the flow of information; the detailed formula is shown in Equation \ref{eq:glu}.
\item \text{KAN}~\cite{liu2024kan} is a neural network based on the Kolmogorov-Arnold representation theorem, which reduces the dependence on linear weight matrices by using learnable functions instead of fixed activation functions.
\item \text{FT-Transformer}~\cite{gorishniy2021fttransfomers} is a model based on the Transformer architecture for the tabular data (Feature Tokenizer + Transformer). It transforms all features (categorical and numerical) into vectors and applies a series of Transformer layers.
\item \text{TabNet}~\cite{arik2019TabNet} is a deep learning model for tabular data that employs attention mechanisms for feature selection, enhancing model performance and interpretability.
\item \text{SAINT}~\cite{somepalli2021saint} is a newly proposed hybrid deep learning approach to solving tabular data problems and performs attention over both rows and columns.
\item \text{NODE}~\cite{Popov2020Node} is a deep learning model that integrates the features of neural networks and decision trees, specifically designed for processing tabular data.
\item \text{DANet}~\cite{danets} is a deep learning framework tailored for tabular data, optimizing feature interaction through Abstract Layers and special shortcut paths.
\end{itemize}

\subsection{Evaluations}
We use KS value and the Area Under the Curve (AUC) to evaluate the results of the credit scoring dataset. The computation of the KS value relies on the correlation between the probabilities or scores predicted by the model and the actual labels. KS value is computed using the formula as 
\(\text{KS = max(TPR - FPR)},\)
where TPR is the True Positive Rate and FPR is the False Positive Rate. The KS value ranges from 0 to 1, and a larger value implies better prediction performance of the model.

AUC is another commonly used evaluation metric for measuring the overall performance of classification models. AUC captures how well the model separates the
two classes in the dataset.  The value of AUC ranges from 0 to 1, where a value closer to 1 indicates stronger classification capability of the model.

\subsection{Implementation Details}
% Our model KMLP consists of one layer of KAN and two layers of gMLP. The hidden size is set to 1024 and the dropout rate is 0.5.
% 
We adopted the Grid Search \cite{grids} technique during the model training process to find the optimal model parameters. We provide them in the Table \ref{tab:params}.
For preprocessing operators of continuous features, the parameter for Quantile Transformation, PLE, and our proposed QTL is the number of bins $n$. On public datasets, given the limited number of features, we set 
$n=100$
 for all three operators. On real-world industrial credit scoring datasets, considering the dimensionality of PLE increases by 
$n$-fold after mapping, 
$n$ for PLE remains set to 100. In contrast, 
$n$ for Quantile Transformation and QTL is set to 1000.
For CLR processing, the calculation follows Equation 10 directly, without the need for any parameter settings.
% 再说明训练参数
In training, the batch size is 4096, the learning rate is initially set to 1e-3 and is decayed by 10\% in every 20 epochs. The optimizer is the Adam optimizer with default configurations. The early stopping technique uses the KS value as the monitor metric for early stopping, with a patience value set to 20.
% 关于其他的模型的参数，我们在附录给出
Regarding the parameters of baseline models, we provide them in the Table \ref{tab:params}.
% 所有的实验结果取10次seed的均值
All of our experiments are repeated with 10 different random seeds, and the average of the experimental results is calculated.
% 训练的设备
Experiments run on a machine equipped with 8 NVIDIA A100-SXM4-80GB GPU and 100 GB RAM , Intel(R) Xeon(R) Platinum 8369B CPU @ 2.90GHz CPU under Ubuntu 20.04 with 64 cores.

\subsection{Model Deployment}
Both QTL and KMLP models are saved in ONNX format, compiled into a user-defined function (UDF) in SQL Server, and deployed for inference. For datasets with over 1 billion samples, the inference process takes approximately 2 hours. Similarly, the LightGBM model is deployed as a UDF in the same manner. During the training phase, a dataset containing 200 million samples is stored in a self-developed file system with high read-throughput. On 8 A100 GPUs, the training time per epoch for the KMLP model is approximately 0.05 hours, with the total training time for 200 epochs summing up to around 10 hours. Due to minor shifts in the temporal feature distributions over time, the model is retrained 1-2 times per year.  
We compared the training time of models on large\-scale dataset(200M), as well as the resource usage and time consumption during online inference after the models were deployed. As shown in the Table \ref{tab:model_deploy}, our KMLP model demonstrates significant advantages in training efficiency compared to LightGBM.

\begin{table}[h]
\centering
\caption{
\small
Comparison of Model Training and Inference.We conducted a comparison of model training and inference in terms of time and resource consumption. The training was performed on a large-scale dataset (200M), while the inference was executed on an ultra-large test set (1B Test). }
\label{tab:model_deploy}
\begin{tabular}{l |c |c}
\hline
\textbf{Model}         & \textbf{KMLP}                    & \textbf{LightGBM}               \\ \hline
Training Resources         & 8 A100 GPUs                     & 10,000 CPUs                     \\ \hline
Training Time              & 10 hours                        & 24 hours                        \\ \hline
Inference Time             & 2 hours                         & 2 hours                         \\ \hline
\end{tabular}
\end{table}

% exp-results
\begin{table}[t]
    \centering
    \caption{
    \small
    Comparison of different numerical preprocessing operators on the medium-scale financial data.}
    \label{tab:num_encoding}
    \begin{tabular}{c c c c|c c}
    \hline 
         \multicolumn{2}{c|}{\multirow{2}{*}{Operator}} & \multicolumn{2}{c|}{Valid} &
         \multicolumn{2}{c}{Test}  \\
         \cline{3-6}
         \multicolumn{2}{c|}{} & AUC & KS & AUC &KS\\ \hline
         \multicolumn{2}{c|}{CLR}    &95.11 &76.34 &94.91 &75.60 \\ 
         \multicolumn{2}{c|}{Quantile} &95.05 &76.51 &94.96 &75.82 \\ 
         \multicolumn{2}{c|}{PLE}      &94.66 &75.89 &94.73 &75.45 \\ 
         \multicolumn{2}{c|}{\textbf{QTL}}      &\textbf{95.20} &\textbf{76.82} &\textbf{95.04} &\textbf{76.08} \\ \hline
    \end{tabular}
\end{table}

\begin{table}[t]
    \centering
    \caption{
    \small
    Experimental results on Medium Scale Financial Data. We highlight the best results in bold, and the second-best results with underscores.}
    \label{tab:ova_results}
    \begin{tabular}{c c c c |c c}
    \hline 
         \multicolumn{2}{c|}{\multirow{2}{*}{Methods}} & \multicolumn{2}{c|}{Valid} &
         \multicolumn{2}{c}{Test}  \\
         \cline{3-6}
         \multicolumn{2}{c|}{} & AUC & KS & AUC & KS\\ \hline
         \multicolumn{2}{c|}{LightGBM} &\underline{ 95.10} &\underline{76.16} &\underline{94.93 }&\underline{75.63} \\ 
         \multicolumn{2}{c|}{MLP}      &94.87 &75.53 &94.71 &75.04 \\ 
         \multicolumn{2}{c|}{gMLP}     &94.99 &75.99 &94.86 &75.41 \\ 
         \multicolumn{2}{c|}{KAN}      &93.15 &73.26 &93.99 &72.86 \\ 
         \multicolumn{2}{c|}{SAINT}    &94.79 &75.73 &94.64 &75.24 \\ 
         \multicolumn{2}{c|}{DANET}    &94.67 &72.28 &94.56 &74.70 \\ 
         \multicolumn{2}{c|}{NODE}     &91.64 &68.63 &91.82 &68.16 \\ 
         \multicolumn{2}{c|}{TabNet}   &94.93 &75.86 &94.80 &75.29 \\ 
         \multicolumn{2}{c|}{FT-Transformer} &95.01 &76.04 &94.92 &75.50 \\  
         \multicolumn{2}{c|}{\textbf{KMLP}}      &\textbf{95.20} &\textbf{76.82} &\textbf{95.04} &\textbf{76.08} \\ \hline
    \end{tabular}
\end{table}

\begin{table*}[t]
    \centering
    \caption{
    \small
    Experimental results on Open Public Datasets. Both AUC and KS metrics indicate that the higher the value, the stronger the model's classification capability; we highlight the best results in bold, and the second best results with underscores.}
    \label{tab:ova_results_pub}
    \begin{tabular}{c c |c c |c c |c c |c c |c c |c c}
    \hline
         \multicolumn{2}{c|}{\multirow{2}{*}{Methods}} & \multicolumn{2}{c|}{CP} &
         \multicolumn{2}{c|}{MT} &
         \multicolumn{2}{c|}{CD} &
         \multicolumn{2}{c|}{EG} &
         \multicolumn{2}{c|}{HI} &
         \multicolumn{2}{c}{JA} 
         \\
         \cline{3-14}
         \multicolumn{2}{c|}{} & AUC & KS &  AUC & KS &  AUC & KS  &  AUC & KS &  AUC & KS  &  AUC & KS \\ \hline
         
         \multicolumn{2}{c|}{LightGBM} &\underline{68.98}	&\textbf{28.27}	&92.24	&69.89	&\underline{84.97}	&\underline{55.30}	&95.05	&75.76	&79.86	&44.16	&86.09	&56.56 \\ 
         \multicolumn{2}{c|}{MLP}      &68.41	&\underline{27.93}	&\underline{93.59}	&72.64	&84.94	&54.88	&\underline{99.4}	&\underline{93.38}	&79.71	&44.00 &\underline{86.27}	&\underline{57.22} \\ 
         \multicolumn{2}{c|}{gMLP}     &68.23	&27.19	&93.35	&\underline{72.66}	&84.75	&54.99	&\textbf{99.56}	&\textbf{95.06}	&79.44	&43.69	&85.62	&56.37 \\ 
         \multicolumn{2}{c|}{KAN}      &67.39	&25.82	&92.10	&68.49	&84.92	&54.91	&98.07	&85.86	&79.32	&43.50
         &86.19	&56.85 \\ 
         \multicolumn{2}{c|}{SAINT}    &67.41	&26.71	&92.10	&69.37	&84.35	&53.82	&98.95	&91.2	&78.36	&41.78	&85.3	&55.05 \\
         \multicolumn{2}{c|}{DANET}    &67.94	&26.91	&93.24	&71.78	&84.91	&54.78	&98.84	&90.29	&79.5	&43.55	&85.95	&56.6 \\
         \multicolumn{2}{c|}{NODE}     &64.73	&22.91	&82.99	&50.58	&83.23	&51.45	&76.25	&41.78	&68.97	&27.33	&80.34	&46.21 \\
         \multicolumn{2}{c|}{TabNet}   &66.4	&24.77	&90.56	&66.31	&83.5	&51.98	&80.42	&46.24	&74.19	&35.81	&80.34	&46.03 \\ 
         \multicolumn{2}{c|}{FT-Transformer} &66.55	&26.59	&92.64	&70.06	&84.95	&55.06	&99.21	&92.16	&\underline{80.84}	&\underline{45.81}	&86.15	&57.03 \\  
         \multicolumn{2}{c|}{\textbf{KMLP}}      &\textbf{69.10}	&27.76	&\textbf{93.72}	&\textbf{73.68}	&\textbf{85.20}	&\textbf{55.93}	&99.16	&91.87 &\textbf{80.88}	&\textbf{46.23}	&\textbf{87.30}	&\textbf{59.51}    \\ 
         \hline
    \end{tabular}
\end{table*}

\begin{figure}[t]
    \centering
\includegraphics[width=\linewidth]{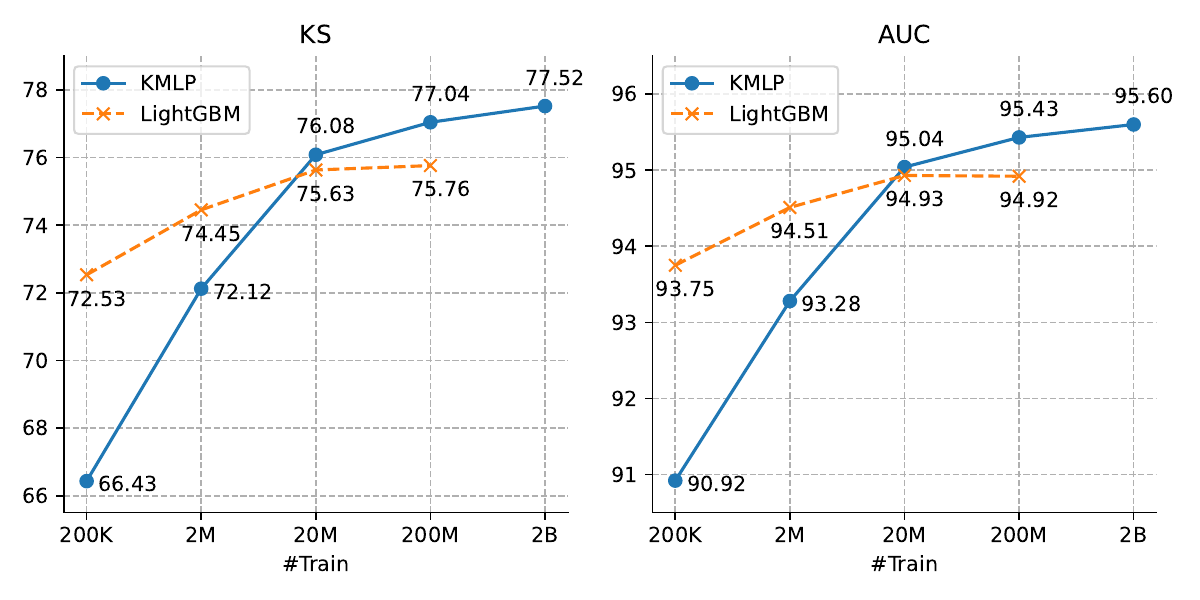}
    \caption{
    \small
    Data Scale Effects. The KS (left) and AUC (right) of LightGBM and KMLP with different sizes of training sets. When the scale of the training data set is relatively small, LightGBM still performs better than KMLP. However, as the scale of the data set increases, KMLP outperformed LightGBM, achieving a 1.76 improvement in the KS value. }
    \label{fig:scaling}
\end{figure}

\begin{table*}[h!]
\centering
\caption{
\small
Ablation Study Performance. The results for the open public datasets are the average across six datasets. The values in parentheses indicate the performance drop compared to the full model, with a downward arrow $\downarrow$ representing the decrease.}
\label{tab:ablation_results}
\begin{tabular}{ l|c|c|c|c}
\hline
\textbf{Model Variant} & \textbf{Average AUC} & \textbf{Average KS} & \textbf{AUC} & \textbf{KS} \\
& \textbf{( Open Public Datasets)} & \textbf{ (Open Public Datasets)} & \textbf{(Industrial Dataset)} & \textbf{(Industrial Dataset)} \\
\hline
\textbf{KMLP (FULL)} & \textbf{73.62} & \textbf{50.71} & \textbf{95.04} & \textbf{76.08} \\
\hline
w.o. QTL & 67.12 ($\downarrow$6.50) & 37.78 ($\downarrow$12.93) & 94.96 ($\downarrow$0.08) & 75.82 ($\downarrow$0.26) \\
\hline
w.o. gMLP & 73.19 ($\downarrow$0.43) & 50.01 ($\downarrow$0.70) & 94.71 ($\downarrow$0.33) & 75.04 ($\downarrow$1.04) \\
\hline
w.o. KAN & 72.99 ($\downarrow$0.63) & 49.99 ($\downarrow$0.72) & 94.86 ($\downarrow$0.18) & 75.41 ($\downarrow$0.67) \\
\hline
\end{tabular}
\end{table*}

\section{Results}
Given the prevalence of a large quantity of continuous data in real-world industrial scenarios, we employed QTL operators to efficiently encode numerical features. Considering the heterogeneous nature of tabular datasets, we proposed a novel model architecture that integrates shallow KAN with deep gMLP. Moreover, recognizing the massive scale of real-world financial data, we systematically investigated the impact of dataset size on the performance of tabular predictive models. In this subsection, we aim to address the following key scientific questions through extensive experiments: 
% \begin{itemize}

 \textbf{(Q1) Preprocessing Capability}: How does the QTL method perform in handling continuous data? What are its advantages compared to other widely-used numerical feature preprocessing techniques?
 
  \textbf{(Q2) Model Performance}: How does the KMLP hybrid architecture perform in tabular data prediction tasks? Does it exhibit competitive performance compared to state-of-the-art tabular prediction models?
  
 \textbf{(Q3) Scalability with Data Size}: As the size of the training dataset increases, how does our model compare to tree-based model LightGBM in terms of performance and generalization ability?
% \end{itemize}

\subsection{Main Results}
\noindent
\textbf{RQ1: Preprocessing Operator Comparison. }
We applied various preprocessing methods to the numerical features of tabular data and fed the processed data into the KMLP model for experiments. The results on public datasets are presented in Table \ref{tab:num_encoding_openml}. The experiments show that our method achieved the best performance on 5 out of 6 datasets. For the medium-sized industrial dataset, the corresponding experimental results are shown in Table \ref{tab:num_encoding}, where QTL also demonstrates superior performance.

Essentially, QTL adopts a transformation similar to quantile normalization, while preserving the linear relationships within each quantile bin. This approach not only standardizes the feature distributions, but also maximally retains the original distribution information, making the data across different feature dimensions more isotropic and preserving important ranking and fine-grained structures. As a result, the model receives more uniform, reasonable, and information-rich inputs, which significantly improves optimization efficiency and generalization capability.

\noindent
\textbf{RQ2: Architecture Results. }
The comparative results of the evaluated architectures are presented in Tables \ref{tab:ova_results_pub} and \ref{tab:ova_results}. Table \ref{tab:ova_results_pub} provides a detailed comparison of the proposed method against other baseline methods on publicly available datasets. Notably, the KMLP architecture exhibited outstanding performance, achieving the highest AUC metric on five out of six datasets. Furthermore, Table \ref{tab:ova_results} showcases the evaluation results on the industrial Medium Scale Financial Dataset. While LightGBM remains a highly competitive approach, our KMLP stands out as the only model that consistently outperformed tree-based models on the test set, setting new benchmarks for state-of-the-art results in both AUC and KS metrics.

\noindent
\textbf{RQ3: Data Scale Effects. }
We evaluated the performance of LightGBM and our proposed model across different training dataset sizes, as shown in Figure \ref{fig:scaling}. On smaller datasets, LightGBM demonstrated significant advantages. For instance, on a dataset with 200K samples, its KS metric outperformed the best-performing neural network method, KMLP, by 6 percentage points. This advantage can likely be attributed to the dataset’s 449 features and high heterogeneity, which allow tree-based models to achieve superior performance when data availability is limited. However, as the dataset size continues to increase, the performance improvements of tree-based models tend to saturate, with limited further gains. In contrast, our proposed model exhibits a consistent upward trend in performance as the dataset size grows. This observation highlights the potential and advantages of our model in leveraging large-scale datasets to achieve superior predictive results compared to tree-based methods.

\subsection{Ablation Anaysis}
% 增加消融实验
To evaluate the contributions of different components in our proposed KMLP, we conducted an ablation study on Open Pubulic and industrial datasets. Specifically, we removed key components including QTL, gMLP , and  the first layer KAN  individually to assess their impact on performance. The results are summarized in Table \ref{tab:ablation_results}, which reports AUC and KS  metrics for each variant.

Specifically, the KMLP (FULL) model achieved superior performance. The noticeable performance degradation upon ablating the QTL module (w.o. QTL) highlights its critical role in preprocessing numerical columns of tabular data. Similarly, the removal of the gMLP module (w.o. gMLP) also led to reduced metric values, indicating its contribution to the model's overall effectiveness in processing tabular information. The absence of the KAN module (w.o. KAN) likewise showed a drop in performance, further emphasizing the positive role each component plays in the KMLP architecture's ability to handle diverse tabular data tasks. These experimental results collectively validate the necessity and individual contributions of each proposed module within the KMLP framework for robust tabular data generation and prediction.

\section{Conclusion}

This paper tackles the challenge of modeling large-scale, dynamic web data, where traditional methods hit scalability limits. We propose KMLP, a novel deep learning architecture that establishes a new paradigm by decoupling feature transformation from high-order interaction modeling. This design automates feature engineering and enables efficient end-to-end learning.
Extensive evaluation on billion-scale industrial data and public benchmarks shows that KMLP not only achieves state-of-the-art performance but, more importantly, exhibits significantly superior scalability compared to GBDTs as data volume and complexity increase. Its successful real-world deployment confirms substantial practical utility and commercial impact.
\begin{acks}
This work was supported by Ant Group Postdoctoral Programme.
\end{acks}

\bibliographystyle{ACM-Reference-Format}
\balance
\bibliography{sample-base}


\clearpage
\appendix
\section{More Dataset Details}
\label{appd:dataset}
 The publicly available dataset we used comes from the OPENML platform. Below, we provide the corresponding data link and statistical description in Table \ref{tab:info} and Table \ref{tab:link}.
% 公开数据集的统计信息以及链接
\begin{table}[h]
    \centering
    \caption{
    \small
    Statistical Information of open public datasets.}
    \label{tab:info}
    \begin{tabular}{l c c c }
    \toprule
    Abbr  & Name &\#samples &\#features     \\ \midrule
    CP    & Click\_prediction\_small & 39948   & 9                \\ 
    MT    & MagicTelescope & 13376   & 10              \\ 
    CD    & Credit &  16714   & 10              \\ 
    EG    & Eeg-eye-state&  14980   & 14              \\ 
    HI    & Higgs & 98050   & 28           \\ 
    JA    & Jannis &  57580   & 54  \\
    \bottomrule
    \end{tabular}
\end{table}

\begin{table}[h]
    \centering
    \caption{
    \small
    The source of the open public datasets.}
    \label{tab:link}
    \begin{tabular}{c c }
    \toprule
    Dataset  & Link     \\ \midrule
    CP    &  https://www.openml.org/d/43901              \\
    MT    &  https://www.openml.org/d/43971              \\ 
    CD    &  https://www.openml.org/d/45024          \\ 
    EG    & https://www.openml.org/d/1471           \\ 
    HI    &  https://www.openml.org/d/42769           \\ 
    JA    &  https://www.openml.org/d/41168   \\
    \bottomrule
    \end{tabular}
\end{table}

% \section{More Details}
% \begin{figure*}
%     \centering
%     \includegraphics[width=\linewidth]{figures/dist2.png}
%     \caption{Simulated Distribution Transformation. Applying the \textbf{QTL} operator to common distributions can lead to a more uniform distribution being obtained and the impact of heterogeneity being eliminated.}
%     \label{fig:QTL-dist}
% \end{figure*}
% %以不同的分布举例说明，我们的算子是怎么变换的
% We present the mapping images of the QTL operator under some classical distributions in Figure \ref{fig:QTL-dist}. The probability density/mass functions of these distributions are as follows.
% \begin{align*}
% &\text{(a)}~\text{Gaussian}(\mu=0,\sigma=1): \quad    p(x) = \frac{1}{\sqrt{ 2 \pi \sigma^2 }}
%                      e^{ - \frac{ (x - \mu)^2 } {2 \sigma^2} }.\\
% &\text{(b)}~\text{Exponential}(\beta=1): \quad p(x) = \frac{1}{\beta} \exp(-\frac{x}{\beta}),\\    
% &\text{(c)}~\text{Beta}(\alpha=0.5,\beta=0.5): \quad    p(x,\alpha,\beta) = \frac{x^{\alpha-1}(1-x)^{\beta-1}}{\mathbf{B}(\alpha,\beta)}.\\
% &\qquad \text{where~} \mathbf{B}(\alpha,\beta) =  \int_0^{1}t^{\alpha-1}(1-t)^{\beta-1}dt.
% \\
% &\text{(d)}~\text{ZIP (Zero-Inflated Poisson Distribution, $\lambda=50$)}:  \\&\qquad  p(x) =
% \begin{cases}
%      \pi+(1-\pi)e^{-\lambda}, ~ &x=0, \\
%     (1-\pi)\frac{\lambda^xe^{-\lambda}}{x!},~ &x>0.
% \end{cases}\\
% &\qquad \text{where,
% $\pi$ is the probability of an excess zero.}
% \end{align*}

% 给出其他的参数的表
\section{Hyperparameters of  EXPERIMENTS}

We use Grid Search to
tune the hyperparameters of our backbone models. The hyperparameter was presented in Table \ref{tab:params}.
For the ultra-large-scale datasets in the industry, we first perform parameter tuning on a smaller dataset (2M) to find the optimal parameters, and then apply these parameters to large or ultra-large datasets.
\begin{table}[t]
    \centering
    \caption{
    \small
    Hyperparameter Space. We use Grid Search to find the optimal parameters. Transfomer-model include FT-Transformer, TabNet and SAINT. }
    \label{tab:params}
\resizebox{\linewidth}{!}{
\tiny
    \begin{tabular}{l l l}
        \toprule
        Models & Parameter & Values\\
        \midrule
        \multirow{6}{*}{LightGBM}   & learning\_rate & \{0.1,0.01\} \\
                                    &  num\_leaves  & \{32, 256,512\}\\
                                    & subsample & \{0.8,0.9,1.0\}\\
                                    & colsample\_bytree & \{0.8, 1.0\} \\
                                    & min\_child\_samples & \{10, 100\} \\
                                    &\#Iterations & \{500,1000, 5000\} \\

        \midrule
        \multirow{4}{*}{NODE} &  \# Layers   & \{2, 4\}  \\
                             & hidden dim & \{1024,2048,3072\} \\
                             & max\_depth & \{6,8\}  \\
                             & tree dim   & 3  \\
        \midrule
       \multirow{2}{*}{DANet}  &  \# Layers   & \{2, 4\}  \\
                                &hidden dim & \{1024,2048,3072\} \\
         \midrule
        \multirow{4}{*}{MLP/gMLP} & \# Layers   & \{2, 4\} \\
                             & hidden dim & \{1024,2048,3072\} \\
                             & learning\_rate & \{0.001,0.005,0.01\} \\
                             & Dropout & \{0.0,0.3,0.5,0.7\} \\
        \midrule
       \multirow{3}{*}{Transfomer-model}   &  \# Layers   & \{2,                                                4\}  \\
                            &hidden dim & \{512,1024,2048,3072\}\\
                            &head & \{4,8\} \\                                     
       \midrule
       \multirow{3}{*}{KAN}  & \# Layers   & \{2, 4\}  \\
                             & grid size & \{5,10\} \\
                             & hidden dim & \{512,1024,2048\} \\
       \midrule
       \multirow{5}{*}{KMLP}  & \# MLP Layers   & \{1, 2\}  \\
                             & \# KAN Layers &  \{1, 2\}\\
                             & grid size & \{5,10\} \\
                             & hidden dim & \{512,1024,2048\} \\
                             & Dropout & \{0.0,0.3,0.5,0.7\} \\
     
       \bottomrule
    \end{tabular}}
    
\end{table}
% \appendix









\end{document}